\documentclass{article}

\usepackage{spconf}
\usepackage{cite}
\usepackage{textcomp}
\usepackage{xcolor}
\usepackage{flushend}

\usepackage{url}
\usepackage{graphicx}%
\usepackage{multirow}%
\usepackage{amsmath,amssymb,amsfonts,amsthm}%
\usepackage{mathtools}
\usepackage{mathrsfs}%
\usepackage{xcolor}%
\usepackage{textcomp}%
\usepackage{manyfoot}%
\usepackage{booktabs}%
\usepackage{algorithm}%
\usepackage{algorithmicx}%
\usepackage{algpseudocode}%
\usepackage{listings}%
\usepackage[acronym]{glossaries}
\usepackage{caption}
\usepackage{subcaption}
\usepackage{cite}
\usepackage{multirow}

\usepackage{booktabs}

\renewcommand{\min}[1]{\underset{#1}{\text{min}}\,}

\newcommand{\argmin}[1]{\underset{#1}{\text{argmin}}\,}
\newcommand{\argmax}[1]{\underset{#1}{\text{argmax}}\,}

\newacronym{ot}{OT}{Optimal Transport}
\newacronym{mmd}{MMD}{Maximum Mean Discrepancy}

\def\BibTeX{{\rm B\kern-.05em{\sc i\kern-.025em b}\kern-.08em
    T\kern-.1667em\lower.7ex\hbox{E}\kern-.125emX}}
\begin{document}

\ninept

\title{Multi-Domain Clustering via Measure Quantization
}

\name{Rafael Pereira Eufrazio$^{1,2}$, Eduardo Fernandes Montesuma$^{3}$ and Charles Casimiro Cavalcante$^{2}$\thanks{This work was partially supported by CNPq Procs. 308512/2023-5 and 420341/2025-0, CNPq/INCT STREAM (Signal processing and TRansmission for Environmental Analysis and Monitoring) 409179/2024-8 and by CAPES - Finance Code 001.}}
\address{$^{1}$Instituto Federal de Educação, Ciência e Tecnologia do Ceará, Canindé-CE, Brazil \\ $^{2}$Federal University of Ceara, Fortaleza-CE, Brazil, \\ $^{3}$Sigma Nova Science, Paris, France}

%
%


\maketitle

\begin{abstract}
\vspace{-0.2 em}
Clustering is a fundamental task in data analysis, typically addressed through centroid-based methods such as K-means. In this work, we present a general framework for multi-domain clustering via measure quantization: given samples from multiple domains, we learn a shared set of cluster prototypes by minimizing a probability metric, such as the Sinkhorn divergence or the Maximum Mean Discrepancy, between each domain's probability measure and the measure of prototypes. Data points are then assigned to clusters either via nearest centroid, or via optimal transport, a collaborative strategy that couples all samples within a domain. A mini-batch optimization strategy makes both fitting and assignment scalable, reducing memory and computational cost while preserving clustering performance. Experimental results on 5 multi-domain benchmarks spanning image, audio and sensor data show that our Sinkhorn-based method consistently outperforms classical and multi-domain clustering baselines, and that this advantage persists when scaling to hundreds of thousands of samples.
\end{abstract}


\section{Introduction}
\vspace{-0.2 em}
Clustering consists of partitioning data-points into groups that share common characteristics. For instance, the well-known K-Means algorithm~\cite{lloyd1982least} partitions the data into groups that share a \emph{centroid}, i.e., a point that aggregates the information of that group. The problem that K-Means solve is a special case of the \emph{Measure Quantization problem}~\cite{cuturi2014fast}, which assumes that data is drawn, i.i.d., from a single probability measure $\mu$. In real applications, data is \emph{heterogeneous}. For instance, in image processing~\cite{patel2015visual}, image data can differ by style, pose, illumination conditions, etc, inducing different statistical properties in the data. This is a known use-case of \emph{Transfer Learning}~\cite{pan2009survey} that develops learning methods (e.g., clustering methods) under the problem of \emph{distribution shift}.

In this paper, we develop a framework for \emph{multi-domain clustering} that generalizes the measure quantization problem to $K$ heterogeneous domains. We assume access to samples $x_{k,i} \overset{\text{i.i.d.}}{\sim} \mu_{k}$, $k=1,\cdots,K$ and $\mu_{k} \neq \mu_{k'} \forall k \neq k'$, and formalize multi-domain clustering as,
\begin{align}
    \nu^{\star} = \argmin{\nu \in \text{Emp}_{C}(\mathbb{R}^{d})}\frac{1}{K}\sum_{k=1}^{K}\mathbb{D}(\mu_{k},\nu),\label{eq:multi-domain-clustering}
\end{align}
where $\mu_{k} \in \text{Emp}_{n_{k}}(\mathbb{R}^{d}) \subset \mathcal{P}(\mathbb{R}^{d})$, $\nu \in \text{Emp}_{C}(\mathbb{R}^{d})$ are measure supported on $C$ points (c.f. equation~\ref{eq:empirical-measures}), and $\mathbb{D}$ is a metric, or notion of dissimilarity between probability measures (e.g., $\mathbb{W}_{2}(\mu,\nu)^{2}$). For $K=1$, equation~\ref{eq:multi-domain-clustering} reduces to the standard measure quantization problem solved by K-Means when $\mathbb{D}=\mathbb{W}_{2}^{2}$~\cite{cuturi2014fast}, and, more generally, is related to the problem of \emph{barycenters of probability measures}~\cite{agueh2011barycenters,cohen2020estimating}. In this perspective, a stream of works~\cite{cuturi2014fast,montesuma2025computing} studies the problem under the Wasserstein distance, a metric between probability measures that comes from \gls{ot} theory~\cite{villani2009optimal,peyre2019computational,montesuma2023recent}. The present work is more general, as it studies problem~\ref{eq:multi-domain-clustering} under different metrics or dissimilarities.

Our contributions are as follows, described where the reader can find them in the manuscript: (i) We propose a new framework for \emph{clustering under distribution shift}, via the measure quantization problem (c.f. equation~\ref{eq:multi-domain-clustering}), by reducing the support of input measures $\{ \mu_k \}_{k=1}^{K}$. (ii) We develop fast, scalable algorithms based on gradient descent of the functional $\nu \mapsto \frac{1}{K}\sum_{k=1}^{K}\mathbb{D}(\mu_{k}, \nu)$ (Section~\ref{sec:mdc-mq}), with respect the centroids $\{z_{1},\cdots,z_{C}\}$ that compose the support of $\nu$. (iii) We experiment with image~\cite{saenko2010adapting,venkateswara2017deep}, audio~\cite{mesaros2018multi} and sensor data~\cite{reinartz2021extended,montesuma2024benchmarking}, showing that our method achieves state-of-the-art performance in comparison with existing clustering baselines (Section~\ref{sec:experiments}).

This paper is organized as follows. Section~\ref{sec:background} includes the background to our paper. Section~\ref{sec:mdc-mq} presents our framework and algorithm. Section~\ref{sec:experiments} shows experiments in multi-domain clustering. Section~\ref{sec:conclusion} concludes this paper.



\section{Background}\label{sec:background}
\vspace{-0.2 em}
\subsection{Clustering}

Let $\{x_i\}_{i=1}^{n} \subset \mathbb{R}^d$ be a dataset drawn from an unknown probability measure $\mu$. A clustering of $\mathcal{X}$ into $C$ groups produces labels $\{y_i\}_{i=1}^{n}$ such that $y_i \in \mathcal{Y}= \{1,\cdots,C\}$. We adopt a \emph{centroid}-based view of clustering, where vectors $\{z_{c}\}_{c=1}^{C}$ represent each cluster $c$ and points are assigned to their nearest centroid, as in the canonical K-Means algorithm~\cite{chong2021k,lloyd1982least}.

One of the main limitations of K-Means is that it fails to cluster data in non-linear structures, such as manifolds. In this sense, other methods exploit the non-linear structure of the data. For instance,~\cite{von2007tutorial} uses the spectral analysis of graph Laplacians to cluster data. In a different direction~\cite{zhang1996birch,ward1963hierarchical} perform clustering based on the construction of a hierarchy within the data. Similarly,~\cite{ho2017multilevel} proposes a multi-level clustering based on the Wasserstein distance.

\subsection{Probability Metrics}

A probability metric is a function $\mathbb{D}:\mathcal{P}(\mathbb{R}^{d})\times\mathcal{P}(\mathbb{R}^{d})\to\mathbb{R}$ that defines a metric over the space of probability metrics, $\mathcal{P}(\Omega)$. In the following, study probability metrics over discrete measures, that is, measures in the set,
\begin{align}
    \text{Emp}_{n}(\mathbb{R}^{d}) = \{ \mu: \mu=\frac{1}{n}\sum_{i=1}^{n}\delta_{x_i}, x_i \in \mathbb{R}^{d} \},\label{eq:empirical-measures}
\end{align}
where $\delta_{x_0}(x) = \delta(x - x_{0})$ is the Dirac measure centered at $x_0$. We study metrics coming from \gls{ot} theory~\cite{villani2009optimal,montesuma2023recent,peyre2019computational} and the \gls{mmd}~\cite{gretton2012kernel}.

In that context, the discrete \gls{ot} problem is given by,
\begin{align}
    T^{\star} = \text{OT}_{\epsilon}(\mu,\nu) = \argmin{\mathbf{T} \in \Pi(\hat{\mu}, \hat{\nu})} \langle T, C \rangle_{F} + \epsilon H(T),\label{eq:OT}
\end{align}
where $C_{ij}$ is called \emph{the ground cost matrix}, $\langle\cdot,\cdot\rangle_{F}$ is the Frobenius inner product, and $H(T)$ is the entropy of the transport plan. $\epsilon = 0$ gives the Kantorovich problem, solvable via linear programming~\cite[Chapter 3]{peyre2019computational}; $\epsilon >0$ gives entropic \gls{ot}, solved by Sinkhorn's algorithm~\cite{cuturi2013sinkhorn}.

When the ground-cost comes from a metric $m$ over $\mathbb{R}^{d}$, that is, $C_{ij} = m(x_{1,i}, x_{2,j})^{p}$, $p \in [1, +\infty)$, the Kantorovich formulation defines a notion of dissimilarity between probability measures,
\begin{align}
    \mathbb{W}_{p,\epsilon}(\mu, \nu)^{p} = \min{\pi \in \Pi(\mu, \nu)}\langle T_{\epsilon}^{\star}, C \rangle_{F} + \epsilon H(T^{\star}_{\epsilon}),
\end{align}
where $T_{\epsilon}^{\star}$ is the \gls{ot} plan. When $\epsilon = 0$, $\mathbb{W}_{p,0}$ is known as the $p-$Wasserstein distance, which is a true metric on $\mathcal{P}_{2}(\mathbb{R}^{d})$. When $\epsilon > 0$, one has the Sinkhorn divergence.

Meanwhile, the \gls{mmd}~\cite{gretton2012kernel} is a kernel-based probability metric. Given a kernel $\kappa:\mathbb{R}^{d}\times\mathbb{R}^{d}\to\mathbb{R}$, it is defined as,
\begin{align}
    \mathbb{MMD}_{\kappa}(\mu,\nu)^{2} &= \dfrac{1}{n^{2}}\sum_{i,j=1}^{n}\kappa(x_{i},x_{j}) + \dfrac{1}{C^{2}}\sum_{c,c'=1}^{C}\kappa(z_{c}, z_{c'}) \notag\\ &- \dfrac{2}{nC}\sum_{i=1}^{n}\sum_{c=1}^{C}\kappa(x_i,z_c).\label{eq:mmd}
\end{align}
In this work, we consider 3 kinds of kernels: linear $\kappa(x,z) = x^{\top}z$, Riesz $\kappa(x,z) = -\lVert x-z \rVert_{2}$ and RBF, $\kappa(x,z) = \exp(-\gamma \lVert x-z \rVert_{2}^{2})$.

\section{Multi-Domain Clustering via Measure Quantization}\label{sec:mdc-mq}
\vspace{-0.2 em}
\begin{figure*}[t]
    \centering
    \includegraphics[width=\textwidth]{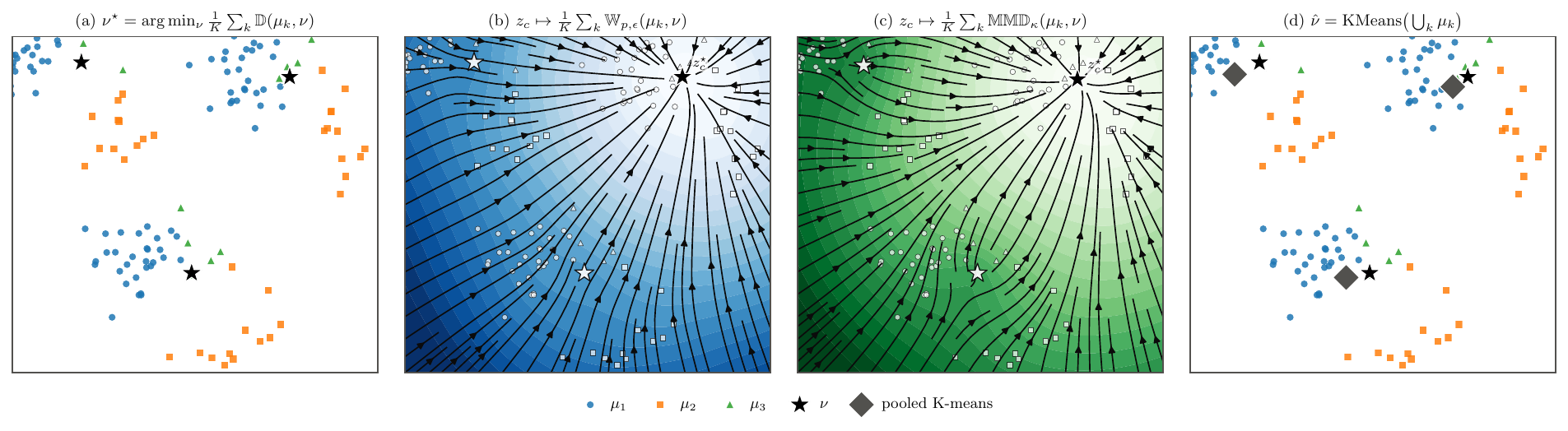}
    \caption{Overview of multi-domain clustering via measure quantization. (a) Toy example with $K{=}3$ domains and the fitted Sinkhorn prototypes $\nu^{\star}$. (b)-(c) Loss landscape $z_c \mapsto \frac{1}{K}\sum_{k}\mathbb{D}(\mu_k,\nu)$ and its gradient flow $-\nabla_{z_c}\frac{1}{K}\sum_{k}\mathbb{D}(\mu_k,\nu)$ (streamlines) for one prototype, while others are fixed (white stars), under $\mathbb{W}_{p,\epsilon}$ and $\mathbb{MMD}_{\kappa}$.}
    \label{fig:methods}
\end{figure*}

In this work, we consider multi-domain clustering. In brief, we have $k=1,\cdots,K$ different domains, each with data $\{x_{i,k}\}_{i=1}^{n_{k}}$. We assume these points are represented in a shared Euclidean space $\mathbb{R}^{p}$. We take a probabilistic view of multi-domain clustering, meaning that we represent each dataset through an empirical probability measure,
\(
    \mu_k = \frac{1}{n_k}\sum_{i=1}^{n_k} \delta_{x_{i,k}}\), and \(\nu = \dfrac{1}{C}\sum_{c=1}^{C}\delta_{z_{c}}
\)
where $\delta_{x_{0}}(x) = \delta(x-x_{0})$ is the Dirac measure centered at $x_0$. Likewise, we represent the distribution of centroids through an empirical measure centered at each centroid. The \emph{multi-domain} clustering task is then finding $Z^{\star} = \{z_{c}^{\star}\}_{c=1}^{C}$ by minimizing the objective in equation~\ref{eq:multi-domain-clustering}, via gradient descent,
\begin{align}
    z_{c,t+1} \leftarrow z_{c,t} - \sum_{k=1}^{K}\lambda_{k}g_{k},
\end{align}
where $g_{k} = \nabla_{z_{c}}\mathbb{D}(\mu_{k},\nu)$ is the gradient of the metric between $\mu_k$ and $\nu$ with respect the support of $\nu$, that is, $z_{1},\cdots,z_{C} \in \mathbb{R}^{d}$. These gradients, with respect the considered metrics, are,
\begin{align}
    g_{k} &= p\sum_{i=1}^{n_{k}}(T_{\epsilon}^{\star})_{ic}m(x_{i,k},z_{c})^{p-1}\nabla_{z_{c}}m(x_{i,k},z_{c}),\label{eq:wass-metric}
\end{align}
and, for the \gls{mmd},
\begin{align}
    g_{k} = \dfrac{2}{C^{2}}\sum_{c=1}^{C}\nabla_{z_{c}}\kappa(z_{c},z_{c'}) - \dfrac{2}{nC}\sum_{i=1}^{n_{k}}\nabla_{z_{c}}\kappa(x_{i,k},z_{c}).\label{eq:mmd-metric}
\end{align}
We summarize these steps in Algorithm~\ref{alg:ot_clustering}, approximating the gradients in equations~\ref{eq:wass-metric}--\ref{eq:mmd-metric} via mini-batching: instead of all $n_{k}$ points, we sample $B \in \mathbb{N}$ points $\{x_{i_j,k}\}_{j=1}^{B}$ per domain, obtaining an estimate $\hat{g}_{k}$.

After obtaining $Z^{\star} = \{z_{1}^{\star},\cdots,z_{C}^{\star}\}$, we assign cluster indices via 2 strategies: \emph{greedy assignment},
\begin{align}
    y_{i,k}^{\star} = \argmin{c \in \mathcal{Y}} m(x_{i,k},z_{c}^{\star}),\label{eq:nn-assign}
\end{align}
i.e., each point to its nearest centroid; and \emph{\gls{ot} assignment},
\begin{align}
    y_{i,k}^{\star} = \argmax{c\in\mathcal{Y}} (T_{\epsilon}^{\star})_{i,c}\text{, where }T_{\epsilon}^{\star} = \text{OT}_{\epsilon}(\mu_k, \nu),\label{eq:ot-assign}
\end{align}
\emph{collaborative} since $T_{\epsilon}^{\star}$ couples all samples $\{x_{i,k}\}_{i=1}^{n_{k}}$ in the support of $\mu_{k}$ (also computable in mini-batches for scalability).

\begin{algorithm}[t]
\caption{Multi-domain Clustering}
\textbf{Input:} Domains $\{\mathcal{X}_k\}_{k=1}^{K}$, number of clusters $C$, 
learning rate $\eta$, number of iterations $N$\\
\textbf{Output:} Cluster prototypes $Z=\{z_c\}_{c=1}^{C}$

\begin{algorithmic}[1]

\State Draw prototypes at random $z_{c}\sim\mathcal{N}(0, \text{Id}), c=1,\cdots,C$

\For{$\tau = 1,\ldots,T$}
    \For{$k = 1,\ldots,K$}
        \State Sample $\{x_{i_{j},c}\}_{j=1}^{B}$, from each $k=1,\cdots,K$
        \State Estimate $\hat{g}_{k} = \nabla_{z_{c}}\mathbb{D}(\mu_{k},\nu)$
    \EndFor
    \State Update prototypes $z_{c,\tau+1} \leftarrow z_{c,\tau} - \eta\dfrac{1}{K}\sum_{k=1}^{K}\hat{g}_{k}$

\EndFor

\State \Return $Z^{\star} = \{z_{c,T}\}_{c=1}^{C}$

\end{algorithmic}
\label{alg:ot_clustering}
\end{algorithm}
\vspace{-0.2 em}

\section{Experiments}\label{sec:experiments}
\vspace{-0.2 em}
In this section, we experiment with multi-domain clustering methods. Our experimentation focuses on classification datasets with multiple domains in 3 areas: computer vision (Office 31~\cite{saenko2010adapting}, Office-Home~\cite{venkateswara2017deep}, Caltech-Office 10~\cite{gong2012geodesic}), audio (TAU Urban Scenes~\cite{mesaros2018multi}) and chemical engineering (Tennessee Eastman Process~\cite{reinartz2021extended,montesuma2024benchmarking}). We additionally experiment with DomainNet~\cite{peng2019moment} for stress-testing the scalability of multi-domain clustering methods. A summary of these datasets is available in Table~\ref{tab:datasets}.

We extract features with ResNet-50~\cite{he2016deep} for Office 31, ResNet-101 for Office-Home, and DeCaf~\cite{donahue2014decaf} for Caltech-Office 10. These networks were trained on ImageNet, and features are extracted without further fine-tuning. For the TAU Urban Scenes dataset, we use the PANN backbone~\cite{kong2020panns}. For the TEP dataset, we follow~\cite{montesuma2025unsupervised} and use 1st and 2nd order statistics of each time series as the features.

\begin{table}[ht]
    \centering
    \resizebox{\linewidth}{!}{
        \begin{tabular}{cccccc}
            \toprule
            Dataset & Modality & \# Features & \# Classes & \# Domains & \# Samples\\
            \midrule
            Office 31 & Image & 2048 & 31 & 3 & 4,110\\
            Caltech-Office 10 & Image & 4096 & 10 & 4 & 2,533\\
            Office-Home & Image & 2048 & 65 & 4 & 15,500\\
            DomainNet & Image & 2048 & 345 & 6 & 586,575\\
            TAU Urban Scenes & Audio & 768 & 10 & 10 & 20,800 \\
            Tennessee Eastman Process & Sensor data & 64 & 29 & 6 & 17,289 \\
            \bottomrule
        \end{tabular}
    }
    \caption{Summary of datasets used in our experiments.}
    \label{tab:datasets}
\end{table}
\vspace{-0.1 em}

We compare 7 methods, grouped into 2 tiers. \emph{Pooled}: 4 classical clustering methods run on the pooled domain data (thus ignoring distribution shift) using Scikit-Learn~\cite{pedregosa2011scikit}. These methods are K-Means~\cite{lloyd1982least}, spectral clustering~\cite{von2007tutorial}, Ward~\cite{ward1963hierarchical} and BIRCH~\cite{zhang1996birch}. \emph{Multi-domain}, which are our Sinkhorn and MMD methods, plus MWMS~\cite{ho2017multilevel}, a multi-level Wasserstein-based clustering mechanism.

We compare these 7 methods through 3 metrics: the Hungarian Accuracy (Hung. Acc.), the Adjusted Rand Index (ARI) and the Normalized Mutual Information (NMI). The Hung. Acc. aligns estimated clusters with the true underlying clusters via an \gls{ot} problem (c.f. equation~\ref{eq:OT}) with uniform marginals. In that case, $T^{\star}_{\text{align}}$ is a permutation matrix~\cite[Chapter 3]{peyre2019computational} and thus makes a correspondence between estimated and underlying clusters. Briefly, the NMI measures how much information the predicted clustering shares with the ground-truth clusters, via a normalized version of mutual information, and the ARI measures the agreement between two cluster predictions by comparing if pairs of samples belong to the same cluster or not. We refer readers to~\cite{vinh10a} for more information about the NMI and ARI metrics.

\begin{table*}[t]\centering
\caption{Per-domain clustering performance (within-domain Hungarian matching, averaged over domains) on the five multi-domain benchmarks. Per dataset: Hungarian Acc.\ / NMI / ARI and their geometric mean (GM); last column: average rank across datasets by GM (lower is better). Sinkhorn uses $p{=}1$, euclidean cost. Best GM per dataset and best avg.\ rank in bold.}
\label{tab:perdomain}
\resizebox{\textwidth}{!}{%
\begin{tabular}{llccccccccccccccccccccc}
\toprule
\multirow{2}{*}{Tier} & \multirow{2}{*}{Method} & \multicolumn{4}{c}{OH} & \multicolumn{4}{c}{O31} & \multicolumn{4}{c}{C10} & \multicolumn{4}{c}{TAU} & \multicolumn{4}{c}{TEP} & \multirow{2}{*}{Avg.\ Rank}\\
\cmidrule(lr){3-6} \cmidrule(lr){7-10} \cmidrule(lr){11-14} \cmidrule(lr){15-18} \cmidrule(lr){19-22}
 & & Acc & NMI & ARI & GM & Acc & NMI & ARI & GM & Acc & NMI & ARI & GM & Acc & NMI & ARI & GM & Acc & NMI & ARI & GM & \\
\midrule
\multirow{4}{*}{Pooled} & KMeans & 0.59 & 0.70 & 0.41 & 0.56 & 0.81 & 0.87 & 0.71 & 0.79 & 0.65 & 0.66 & 0.43 & 0.57 & 0.42 & 0.38 & 0.24 & 0.34 & 0.24 & 0.44 & 0.05 & 0.18 & 4.6\\
 & Spectral & 0.58 & 0.70 & 0.33 & 0.51 & 0.73 & 0.84 & 0.55 & 0.70 & 0.62 & 0.69 & 0.38 & 0.55 & 0.40 & 0.41 & 0.22 & 0.33 & 0.30 & 0.60 & 0.16 & 0.31 & 5.4\\
 & Ward & 0.60 & 0.70 & 0.39 & 0.55 & 0.85 & 0.89 & 0.75 & \textbf{0.83} & 0.67 & 0.72 & 0.50 & 0.62 & 0.40 & 0.35 & 0.22 & 0.32 & 0.27 & 0.48 & 0.06 & 0.19 & 4.2\\
 & BIRCH & 0.59 & 0.70 & 0.41 & 0.56 & 0.84 & 0.89 & 0.73 & 0.82 & 0.66 & 0.69 & 0.50 & 0.61 & 0.40 & 0.35 & 0.22 & 0.32 & 0.21 & 0.40 & 0.04 & 0.15 & 4.6\\
\midrule
\multirow{3}{*}{Multi-domain} & MWMS & 0.59 & 0.69 & 0.46 & 0.57 & 0.79 & 0.84 & 0.70 & 0.77 & 0.72 & 0.71 & 0.51 & 0.64 & 0.42 & 0.38 & 0.24 & 0.34 & 0.22 & 0.50 & 0.08 & 0.21 & 3.4\\
 & Sinkhorn (ours) & 0.63 & 0.72 & 0.52 & \textbf{0.62} & 0.81 & 0.86 & 0.74 & 0.80 & 0.85 & 0.80 & 0.72 & \textbf{0.79} & 0.43 & 0.38 & 0.24 & \textbf{0.34} & 0.63 & 0.71 & 0.58 & \textbf{0.64} & \textbf{1.4}\\
 & MMD (ours) & 0.56 & 0.66 & 0.39 & 0.52 & 0.82 & 0.85 & 0.72 & 0.79 & 0.70 & 0.66 & 0.49 & 0.61 & 0.41 & 0.36 & 0.22 & 0.32 & 0.29 & 0.52 & 0.11 & 0.26 & 4.4\\
\bottomrule
\end{tabular}}
\end{table*}
\vspace{-0.3 em}

\begin{figure}[ht]
    \centering
    \includegraphics[width=\linewidth]{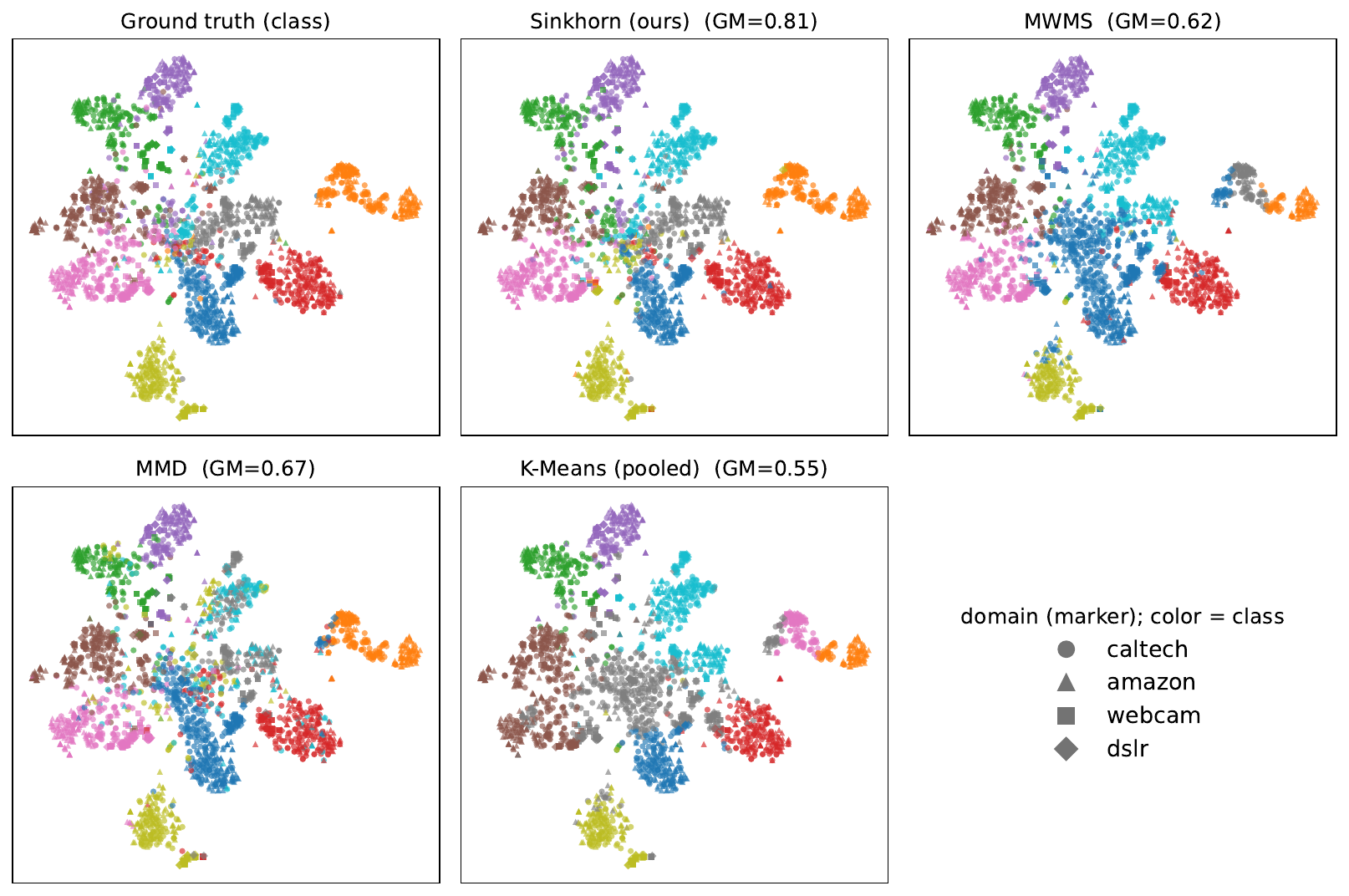}
    \caption{Clustering methods over the Caltech-Office 10 dataset. Overall, using the Sinkhorn divergence as a criterion yields the most consistent clustering of the data.}
    \label{fig:tsne-vis}
\end{figure}
\vspace{-0.1 em}

All experiments ran on a single machine (AMD EPYC 7413 CPU, NVIDIA L4 GPU, 47\,GB RAM) using PyTorch~\cite{paszke2019pytorch} and PythonOT~\cite{flamary2021pot}. Our results are summarized in Table~\ref{tab:perdomain}. Overall, methods that exploit the multi-domain structure of the data rank best than \emph{pooled}, classical clustering algorithms. Out of the 3 tested multi-domain methods, those using the Wasserstein geometry (ours, and MWMS~\cite{ho2017multilevel}) have the best performance. Using the \gls{mmd} achieves an average rank comparable to MWMS. For instance, we visualize in Figure~\ref{fig:tsne-vis} the clustering of each method on the Caltech-Office 10 dataset~\cite{gong2012geodesic}. Next, we ablate our method across 2 angles.

\begin{figure}[ht]
    \centering
    \includegraphics[width=\linewidth]{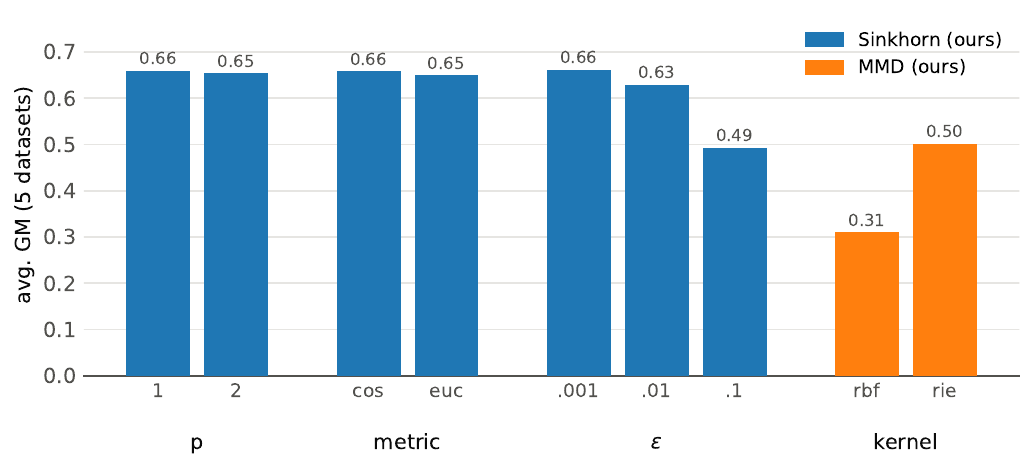}
    \caption{Ablation of hyper-parameters $(p, m, \epsilon, \kappa)$ of our methods.}
    \label{fig:ablation}
\end{figure}
\vspace{-0.3 em}

On the first angle, we ablate the Sinkhorn divergence $\mathbb{W}_{p,\epsilon}$ over $p \in \{1,2\}$, $\epsilon \in \{10^{-3},10^{-2},10^{-1}\}$, $m \in \{\text{Cos},\text{Eucl}\}$, and the \gls{mmd} kernel (linear, RBF, Riesz), reporting for each value its best configuration over the remaining hyper-parameters, averaged across the 5 datasets. $p$ and $m$ have little impact on GM ($\leq 0.01$ apart), whereas $\epsilon$ does: GM drops from $0.66$ at $\epsilon{=}10^{-3}$ to $0.49$ at $\epsilon{=}10^{-1}$, as heavier entropic smoothing blurs the coupling between samples and prototypes. The \gls{mmd} kernel matters even more: switching from RBF to a Riesz kernel raises GM by $0.19$ ($0.31 \to 0.50$), making kernel choice the dominant lever for \gls{mmd}-based clustering.

\begin{figure}[ht]
    \centering
    \includegraphics[width=\linewidth]{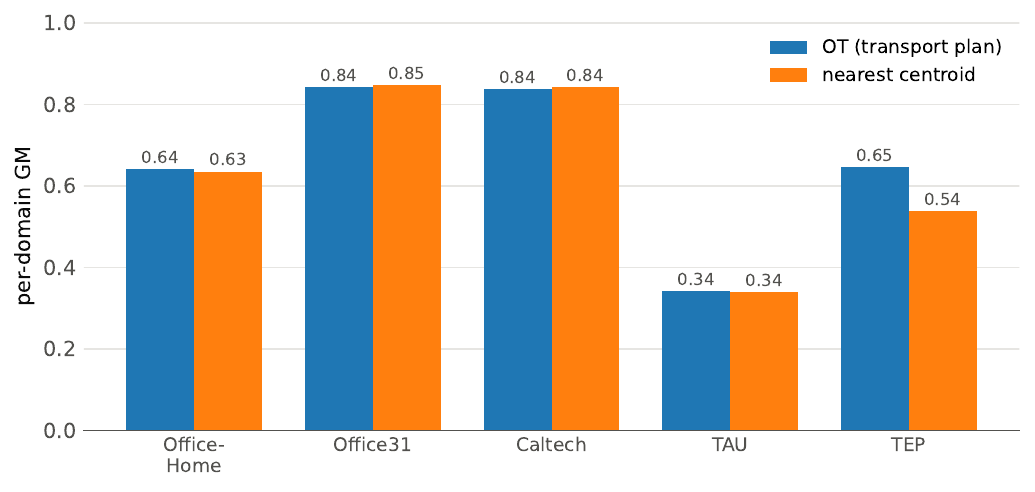}
    \caption{Ablation of assignment mechanism of our methods.}
    \label{fig:assign_ablation}
\end{figure}
\vspace{-0.3 em}

On the second angle, we compare \gls{ot} assignment (equation~\ref{eq:ot-assign}) against nearest centroid (equation~\ref{eq:nn-assign}) for the best Sinkhorn configuration per dataset. The two strategies are within $0.01$ GM of each other on Office-Home, Office31, Caltech-Office10 and TAU, but \gls{ot} assignment gives a $+0.11$ GM gain on TEP ($0.65$ vs.\ $0.54$), the dataset with the most classes (29) and the most severe class imbalance, suggesting the collaborative, transport-plan-based assignment helps most when clusters are hard to disambiguate from a single centroid alone.

\begin{figure}[ht]
    \centering
    \includegraphics[width=\linewidth]{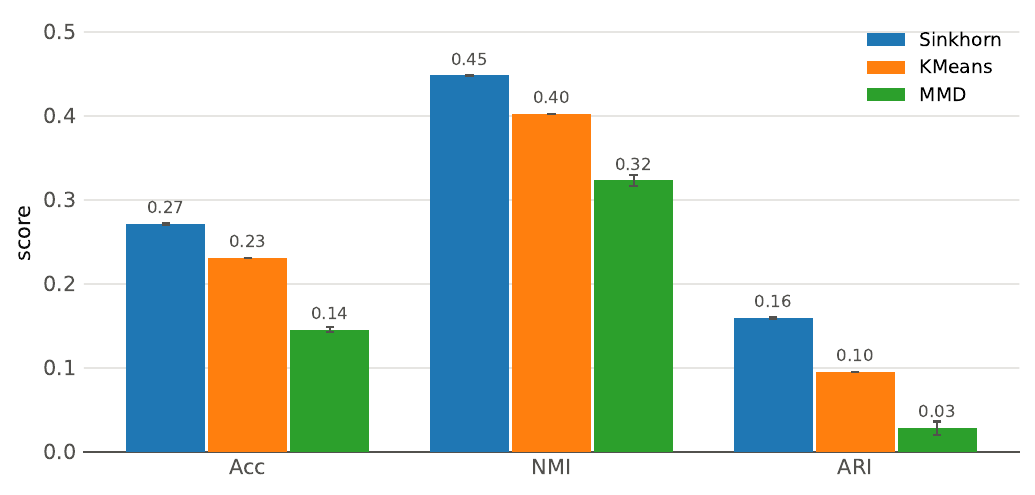}
    \caption{Scaling experiment on DomainNet.}
    \label{fig:DomainNet}
\end{figure}
\vspace{-0.3 em}

Finally, we stress-test scalability on DomainNet (586,575 samples, 6 domains, 345 classes), comparing our methods against mini-batch K-Means (batch size 8,192). Sinkhorn (ours) outperforms mini-batch K-Means on all 3 metrics (Acc: $0.27$ vs.\ $0.23$; NMI: $0.45$ vs.\ $0.40$; ARI: $0.16$ vs.\ $0.10$), while \gls{mmd} (ours) trails both ($0.14$/$0.32$/$0.03$). Overall, combining these results with those in Table~\ref{tab:perdomain}, we conclude that the Wasserstein distance yields a richer geometry for clustering the underlying measures. This confirms that our Sinkhorn-based approach retains its advantage over classical clustering even with hundreds of thousands of samples.

\section{Conclusion}\label{sec:conclusion}
\vspace{-0.2 em}
In this work, we propose a general framework for multi-domain clustering via measure quantization, i.e., the reduction of a set of measures' support to $C$-prototypes by gradient descent of probability metrics~\cite{montesuma2025computing}, scalable via mini-batching~\cite{fatras2021minibatch}. Our experiments demonstrate the superiority of the proposed framework \emph{when a suitable geometry in the space of probability measures is defined} (e.g., the Sinkhorn divergence~\cite{genevay2018learning}). Future work includes the theoretical study of the convergence of our algorithm, and consideration of other geometries, especially when domains live in incomparable spaces (e.g., via the Gromov-Wasserstein metric~\cite{memoli2011gromov}).

\section{Compliance with Ethical Standards}
\vspace{-0.2 em}
This computational study used publicly available benchmark datasets
and did not involve the recruitment or intervention of human or animal
subjects. Therefore, ethical approval was not required.

\section{Acknowledgments}
\vspace{-0.2 em}
The authors declare that they have no relevant financial or
nonfinancial interests to disclose.

\bibliographystyle{IEEEbib}
\bibliography{refs}

\end{document}